\documentclass[conference]{IEEEtran}

\usepackage{cite}
\usepackage{amsmath,amssymb,amsfonts}
\usepackage{amsthm}
\usepackage{graphicx}

\usepackage{algorithm}
\usepackage{setspace} 
\usepackage{algpseudocode} 
\usepackage{cite}
\usepackage[bookmarks=false]{hyperref}
\usepackage{subfigure}

\usepackage [font=scriptsize]{caption}
\usepackage{url}

\def\BibTeX{{\rm B\kern-.05em{\sc i\kern-.025em b}\kern-.08em
    T\kern-.1667em\lower.7ex\hbox{E}\kern-.125emX}}

\theoremstyle{definition}

\begin{document}
%

\title{Learning Obstacle-Avoiding Lattice Paths using Swarm Heuristics: Exploring the Bijection to Ordered Trees}


\author{{Victor Parque}\\
\IEEEauthorblockA{ \emph{ Department of Modern Mechanical Engineering, Waseda University} \\
3-4-1 Okubu Shinjuku Tokyo, 169-8555, Japan\\
parque@aoni.waseda.jp}
}



\IEEEoverridecommandlockouts

\maketitle

\IEEEpubidadjcol

\begin{abstract}
Lattice paths are functional entities that model efficient navigation in discrete/grid maps. This paper presents a new scheme to generate collision-free lattice paths with utmost efficiency using the bijective property to rooted ordered trees, rendering a one-dimensional search problem. Our computational studies using ten state-of-the-art and relevant nature-inspired swarm heuristics in navigation scenarios with obstacles with convex and non-convex geometry show the practical feasibility and efficiency in rendering collision-free lattice paths. We believe our scheme may find use in devising fast algorithms for planning and combinatorial optimization in discrete maps.
\end{abstract}




%
\IEEEpeerreviewmaketitle

\begin{IEEEkeywords}
path planning, lattice paths, ordered trees, enumeration, combinatorial objects, Catalan numbers
\end{IEEEkeywords}

\section{Introduction}

Trees allow to model hierarchical dependencies in combinatorial problems ubiquitously\cite{aldana20}. Among the existing class of trees, the family of ordered trees are well-known to have a bijection to Catalan numbers, binary trees with $n$ external nodes, trees with $n$ nodes, legal sequences of $n$ pairs of parentheses, triangulated $n$-gons, and lattice paths\cite{kreher98}.

How to generate ordered trees has attracted the favorable attention in the discrete mathematics community. Among the existing approaches, integer sequences are known to generate ordered trees with $n$ vertices and $k$ leaves in $O(n-k)$ time\cite{pallo87}. Also, it is possible to traverse the \emph{genealogy of trees} with at most $n$ vertices in $O(n)$ space and $O(1)$ time per tree in average\cite{naka02}. The \emph{genealogy of trees} is in essence the tree of trees, and offers a systematic means to render all families of trees. For instance, it is possible to enumerate ordered trees with exactly $n$ vertices and $k$ leaves in $O(1)$ time in the worst case\cite{yaka09}. Most of the above-mentioned approaches basically extend the notion of reverse search\cite{reverse96}, which generates objects through a graph (tree) whose edges model local and bounded operations on the objects, thus it becomes possible to generate entities (trees) by traversing the graph backwards by an adjacency expansion oracle. Authors have also explored the idea of complete generation of trees\cite{knuth06}, and others have used distinct mechanisms such as parent arrays\cite{liruskey99,imcom21}, level sequences\cite{sawa06}, structural constraints\cite{sen17}, binary trees\cite{at92}, triangulations of convex polygons\cite{jing16} and lattice enumerations\cite{hislattice10}.

In this paper we tackle the problem of planning obstacle-avoiding paths over grid maps by using gradient-free heuristics and a representation rendered from the bijection between lattice paths and the class of ordered trees. Generally speaking, path planning by heuristic algorithms are often based on A*\cite{astar18}, Genetic Algorithms\cite{ga21,gacec21}, Differential Evolution\cite{de17,acc20}, Heuristic Graph Search\cite{oli18}, Ant Colony Optimization\cite{acogecco20}, Particle Swarm Optimization\cite{psocec16} and Local Search\cite{compsac20,case21}. Basically, the path planning problem has been approached from the mutation, crossover and hybrid selection perspectives, in which paths are often represented by a set of points or control commands whose optimal configuration is to be found by the optimization heuristic. Thus, the problem scales with the number of points or commands used in the encoding.

On the other hand, we exploit the 1-1 correspondence between ordered trees and the space of lattice paths\cite{imcom21} to devise an heuristic to generate obstacle-avoiding lattice paths, whose unique benefit is to render a one-dimensional search problem, which is tractable by the state of the art search/optimization heuristics. The study of the path planning problem using the bijection to the ordered trees and its feasibility study by swarm optimization algorithms is the first proposed in the literature, to the best of our knowledge. Basically, we present a new approach to generate collision-free lattice paths in grid maps by using a bounded search space over the combinatorial encoding of ordered trees. Our contributions are summarized as follows:

\begin{itemize}
\item a recursive approach that generates obstacle-avoiding lattice paths over a one-dimensional search space.
\item the computational studies using relevant nature-inspired swarm heuristics using diverse set of exploration-exploitation features in convex and non-convex navigation scenarios.
\end{itemize}

Our experiments demonstrated that Strategy Adaptation Differential Evolution (SADE) and the swarm algorithms showing the explorative features outperformed their counterparts with exploitative features. These observations suggest that the swarm-based algorithms using explicit exploration mechanisms offer potential merits to tackle the lattice path planning problem using the ordered tree representation effectively.

\section{Proposed Method}

In this section, we present the main concepts and algorithms involved in our proposed approach.

\subsection{Encoding Mechanism}

\begin{figure}[t!]
	\centering
	\includegraphics[width=0.98\columnwidth]{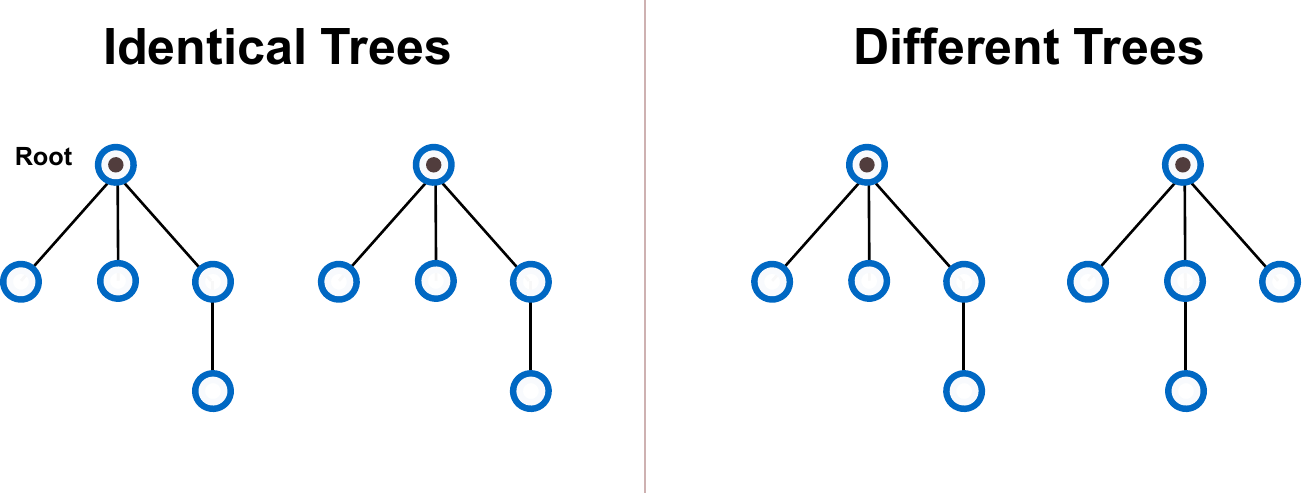}
	\caption{Main notion of ordered trees}
	\label{otree}
\end{figure}

Due to the nature of the implicit order of leaves, two ordered trees are similar (different) if the order of leaves follow the same (different) order with respect to some convention, as the one from left to right as shown by Fig. \ref{otree}. The nature of such order allows one to use a encodings based on tree traversal such as the pre-order or post-order arrangements to identify nodes in the tree systematically. In this paper, we use the following tuple-based mechanism to represent an ordered tree:

\begin{equation}\label{t}
t = (t_1, t_2, ..., t_i, ...,  t_n),
\end{equation}
where an ordered rooted tree has $n$ nodes and $t_i$ stands for the number of children of the $i$-th node of the tree in preorder traversal, thus

\begin{equation}\label{condt}
t_n = 0, ~ t_i \in [0, n-1].
\end{equation}

\begin{figure}[t!]
	\centering
	\includegraphics[width=0.98\columnwidth]{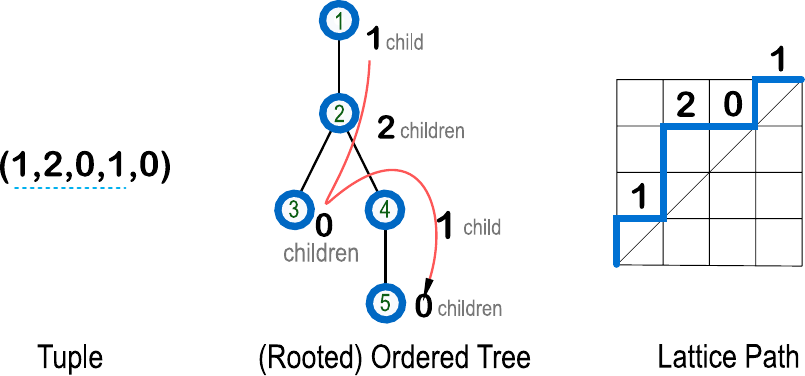}
	\caption{Basic idea of the encoding mechanism. Each element of the tuple denotes the number of children of the ordered tree in preorder labeling of the nodes. Also, each element of the tuple denotes the relative height of the lattice path.}
	\label{basic}
\end{figure}

The above-mentioned representation in (\ref{t}) has a 1-1 bijection to the family of lattice paths in a grid with $n \times n$ nodes as shown by the example of Fig. \ref{basic}. In such case, $t_i$ denotes the relative height in the grid.

Also the above tuple-based representation is inspired by the BCT representation of binary trees, in which B stands for branching, C stands for continuation, and T stands for terminal\cite{bct,eeckman94}.

\begin{itemize}
  \item The BCT representation is renderable from the column-wise sum of elements below the diagonal of the adjacency matrix representing the tree, and
  \item The adjacency matrix is renderable from the BCT encoding by an $O(n)$ algorithm based on stacks\cite{cuntz10d}.
\end{itemize}

Under the BCT approach, if ordered trees were represented by an adjacency matrix, and assuming tree nodes are labeled by a user-defined order, each element of the tuple $t$ in (\ref{t}) is equivalent to the column-wise sum of elements below the adjacency matrix representation.

\begin{figure*}[t!]
	\centering
	\includegraphics[width=0.95\textwidth]{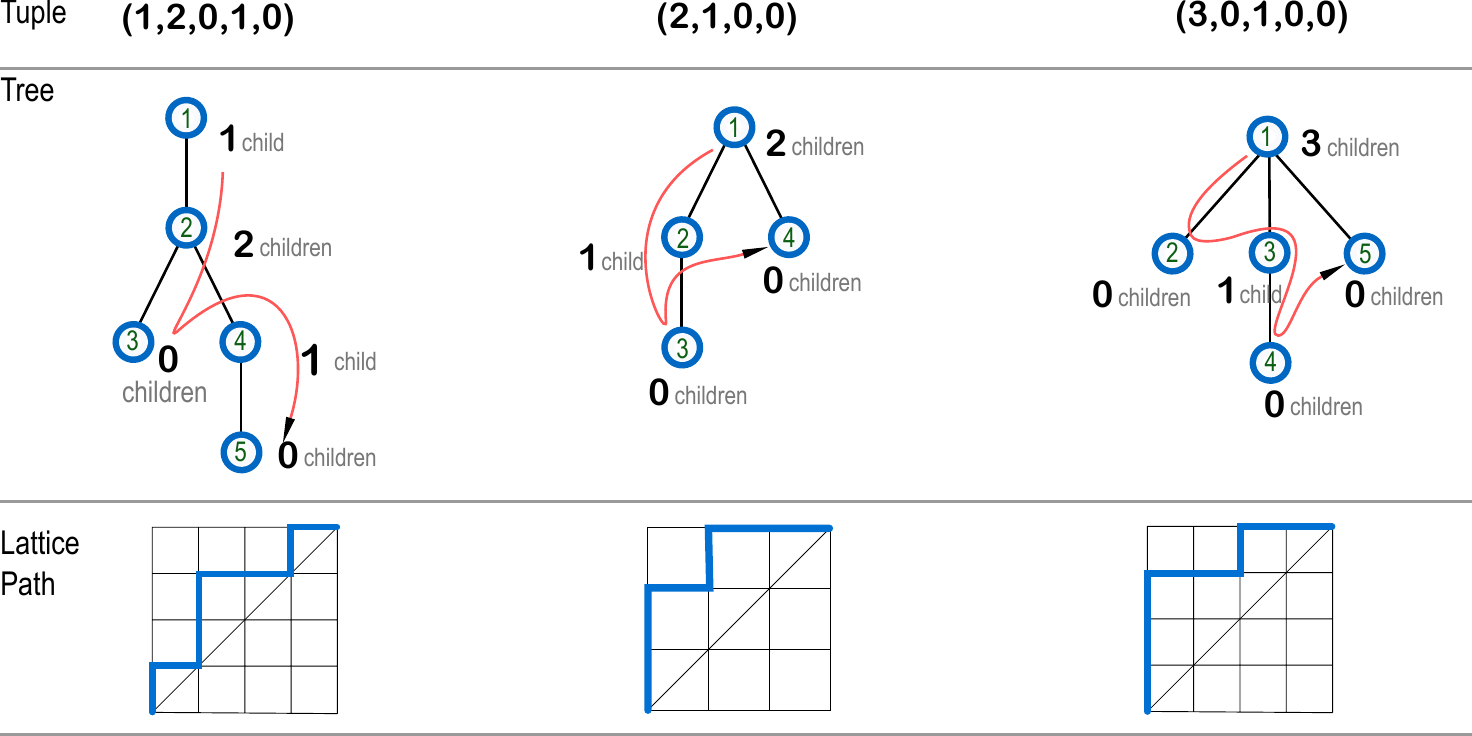}
	\caption{Examples of the bijection between lattice paths and ordered trees. Elements of the tuple denoted above the picture denote the number of children of each node in the tree in preorder encoding. To represent the lattice path, the digits of the tuple encode the relative column height.}
	\label{enco}
\end{figure*}

\begin{figure}[ht!]
	\centering
	\includegraphics[width=0.98\columnwidth]{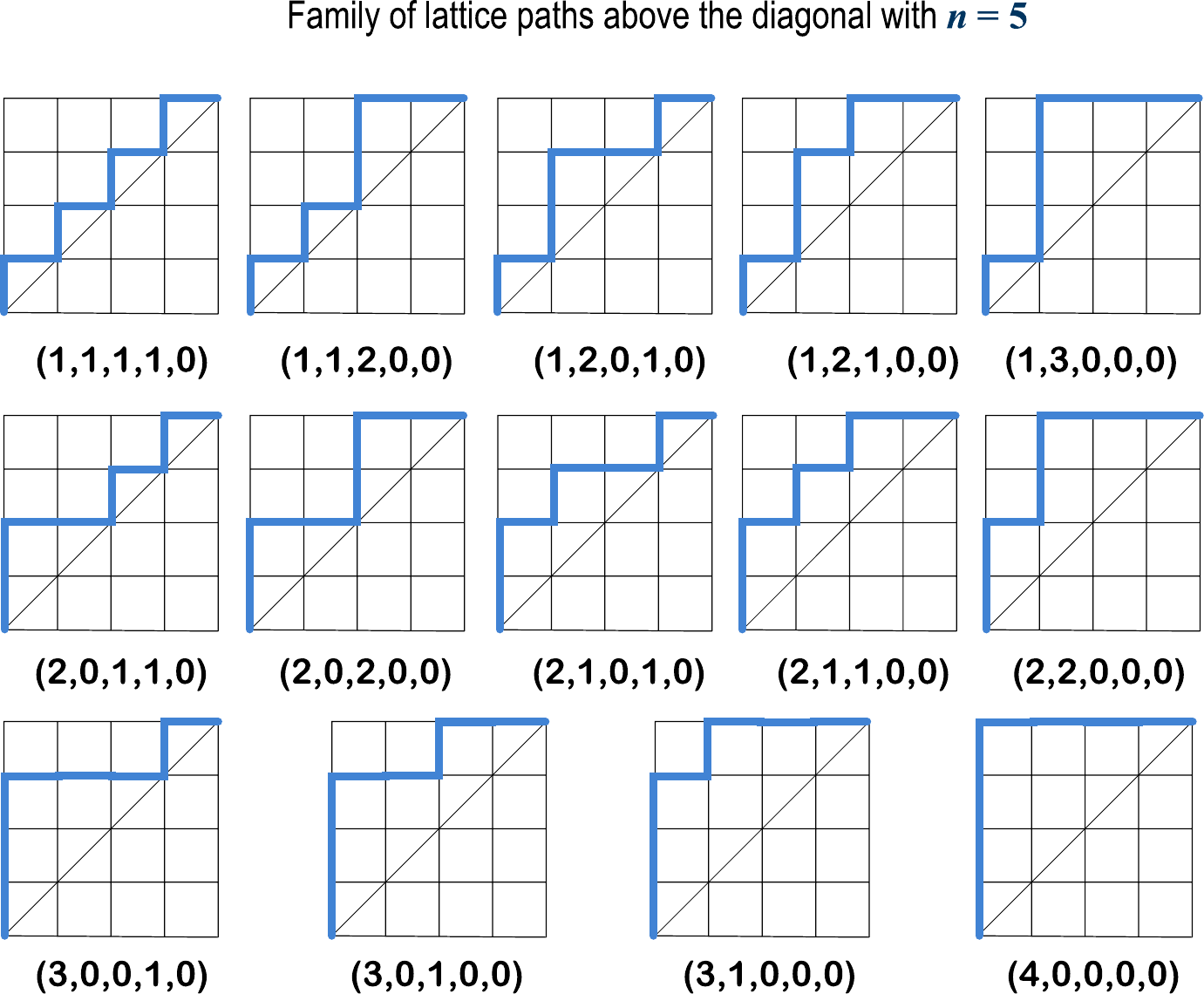}
	\caption{The 1-1 correspondence with the lattice paths above the diagonal of a grid with $(n-1) \times (n-1)$ square cells for $n=5$. }
	\label{lattice5}
\end{figure}

To give an example of the 1-1 bijective relationship between ordered trees and lattice paths, Fig. \ref{enco} shows the node labels that define the traversal order in pre-order and their 1-1 parallel to lattice paths; and Fig. \ref{lattice5} shows examples of the encoding of diverse lattice paths of a square grid of $n = 5$.

Furthermore, due to the tuple $t$ encodes the relative height of the node in the $n \times n$ lattice, the following holds:

\begin{equation}\label{sumbct}
\sum_{i = 1}^{n} t_i = n-1
\end{equation}

The above can also be derived from the following notion: since $t_i$ is equivalent to the number of edges of the $i$-th node, then summing up all elements of the tuple $t$ will render the total number of edges of the tree, that is $n-1$.

\subsection{Generating Lattice Paths}

By using the tuple-based representation in (\ref{t}), we propose a mechanism to generate arbitrary lattice paths which are above the diagonal. As such, it is possible to generate lattice paths by finding each element $t_i$ of the tuple $t$ for $i \in [n]$ by the following relationships:

\begin{equation}\label{tisample}
t_i \sim  \mathbf{U}\{L_i, U_i\} ~ i = 1, 2, ..., n
\end{equation}

\begin{equation}\label{Li}
L_i = 1 - \text{sgn} \Big (  t_{i-1} + S_{i-1} - 1 \Big )
\end{equation}

\begin{equation}\label{Si}
S_i = S_{i-1} + t_{i-1} - 1
\end{equation}

\begin{equation}\label{Ui}
U_i = U_{i-1} - t_{i-1}
\end{equation}
where $L_i$ and $U_i$ are the lower and upper bound on $t_i$, respectively, such that $t_i \in [L_i , U_i]$, and sgn(.) denotes the signum function. Since the above mechanism is recursive in nature, we set the initial conditions $S_0 = 0$, $U_0 = n$, $t_0 = 1$. The variable $S$ computes the accumulation of elements of the tuple. It is possible to eliminate the variable $S$, by which an equivalent expression to Eq. \ref{Ui} is

\begin{equation}\label{Uiv2}
L_i = 1 - \text{sgn} \Bigg (  \sum_{j = 0}^{i-1} (t_j - 1) \Bigg )
\end{equation}

For ordered trees with $n$ nodes, and considering the ordered nature of the tuple $t$ and of Eq. \ref{Li} and Eq. \ref{Ui}, the following relations hold: $L_1 = 1$, $U_1 = n-1$, and $L_n = U_n = 0$, thus $t_n = 0$, which aligns well with Eq. \ref{condt}, and

\begin{equation}\label{t1}
t_1 \in [1, n-1].
\end{equation}

\subsection{Sampling Lattice Paths}

Since the bounds for $t_i$ are given as

\begin{equation}\label{tibound}
t_i \in [L_i \ldots U_i],
\end{equation}
it becomes possible to compute the tuple $t$ stochastically to realize the sampling mechanism of (\ref{tisample}) as follows:


\begin{equation}\label{tipath}
t_i  =  \Bigg \lfloor L_i +\lambda( U_i - L_i ) \Bigg \rceil,
\end{equation}

\begin{equation}\label{lambda}
\lambda = \mathbf{r} . \Bigg ( \frac{x_i}{n} \Bigg ) ^\alpha,
\end{equation}
where $\lambda$ is a normalization factor, $\mathbf{r}$ is a random number with uniform distribution $\mathbf{U}[0, 1]$, $\alpha$ is curvature preference, $x_i$ is the x-coordinate of the $i$-th node of the lattice path, for $i = 1, 2, ..., n,$. The range of $\alpha \in [0, \infty]$: small values of $\alpha$ (close to zero) denote $L$-shaped lattices paths (orthogonal to both x-axis and y-axis), and the large values of $\alpha$ denote lattice paths being close to the diagonal.

\begin{figure}[t!]
	\centering
	\includegraphics[width=0.65\columnwidth]{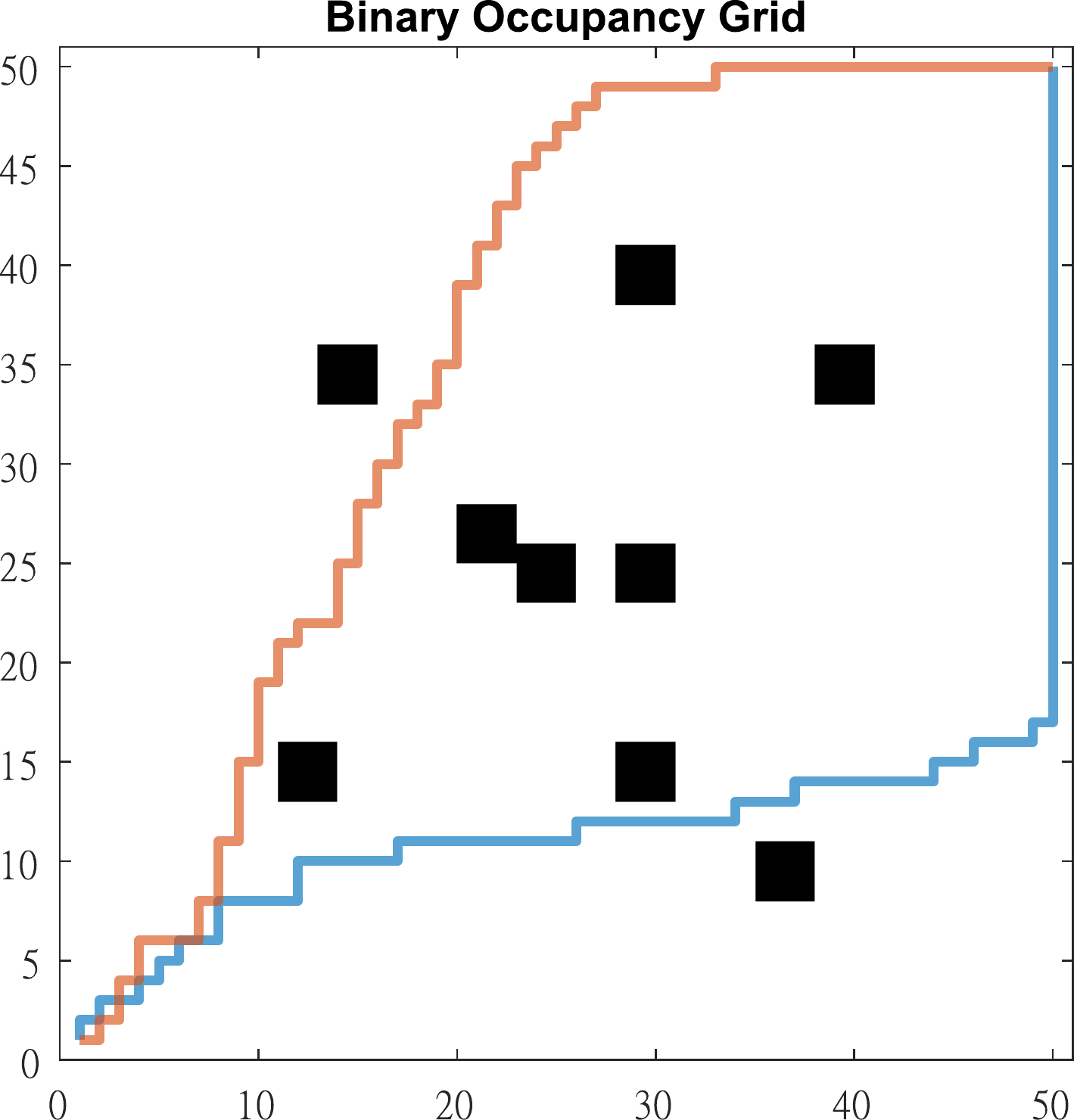}
	\caption{Example of obstacle-avoiding lattice paths above and below the diagonal.}
	\label{example}
\end{figure}

\begin{algorithm}[t!]
\caption{Generate Lattice Path}\label{gen}
\begin{spacing}{1.1}
\begin{algorithmic}[1]
\Function{Generate Path}{$L, U, S, n, \alpha, \lambda, x, y$}

\If{$n > 1$ }

\State Compute $t = \Big \lfloor L +\lambda( U - L ) \Big \rceil   $

\If{$x = 1$ }
\State $y = t$
\Else
\State $y = y + t$
\EndIf

\State $O_{x, y} \gets$ Get Occupancy at $(x, y)$

\If{$O_{x, y}  =  $  {False}}
\State $u \gets U - t$
\State $S \gets S + t - 1$
\State $l \gets 1 - \text{sgn}(S)$
\State $ \lambda = \mathbf{r} . \Big ( \displaystyle \frac{x}{n} \Big ) ^\alpha$
\State $ n \gets n - 1$
\State $t' \gets$ \Call{Generate Path}{$l, u, S, n, \alpha, \lambda, x+1, y$}
\State \Return $t \cup t'$

\Else
\State \Return $\{ \}$
\EndIf
\State \Return $\{ \}$

\EndIf

\EndFunction
\end{algorithmic}
\end{spacing}
\end{algorithm}

\begin{figure*}[t!]
	\centering
	\includegraphics[width=0.98\textwidth]{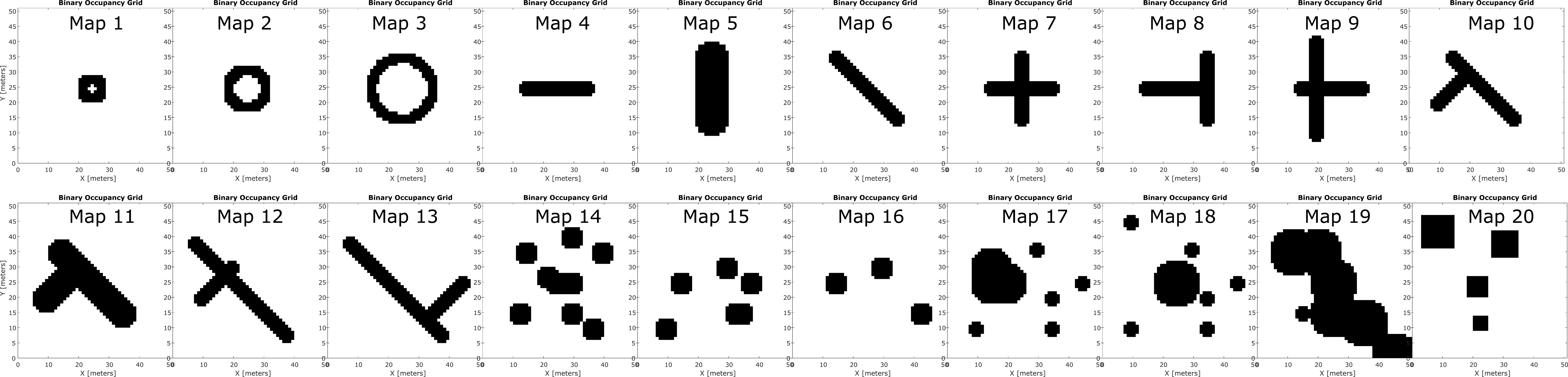}
	\caption{Grid map environments used for evaluation.}
	\label{maps}
\end{figure*}

Computing (\ref{Li}) - (\ref{Ui}) implies a recursive procedure with $\mathcal{O}(n)$ space and $\mathcal{O}(1)$ time in average per path, realized by algorithm \ref{gen} generating lattice paths above the diagonal. Here, inputs are: the lower bound  $L$ of the $i$th element of the tuple $t$, the upper bound $U$ of the $i$th element of the tuple $t$, the summation term $S$, the number of nodes $n$ in the grid, the preference for curvature $\alpha$, the normalization factor $\lambda$, the initial coordinates in the grid $(x, y)$. Note that line 9 of algorithm \ref{gen} checks the occupancy at coordinate $(x, y)$, thus it is possible to compute collision-free paths above the diagonal.

A lattice path starting at (0, 0) and ending at the upper corner of the grid is generated by algorithm \ref{gen} with inputs $L= 1, U = n-1, S = 0, \lambda = 0, x = 0, y = 0$ and the user-defined parameter $\alpha$. Although algorithm \ref{gen} generates a path above the diagonal, it is possible to swap the coordinate $x$ by the coordinate $y$ to generate paths below the diagonal. For instance, Fig. \ref{example} shows two lattice paths, one above the diagonal (with $\alpha = 1.2$), and another below the diagonal (with $\alpha = 2.5$), implying that the parallel execution of algorithm \ref{gen} allows to compute multiple collision-free paths above and below the diagonal, for which the path topology depends on the one-dimensional parameter $\alpha$. In Fig. \ref{example}, the origin is located at the bottom-left, while the destination is located at the top-right corner. In the following section, we evaluate the possibility of using gradient-free optimization heuristics to find optimal values of $\alpha$ for distinct navigation conditions.


\section{Computational Experiments}

In order to evaluate the feasibility of generating collision-free paths, we performed computational experiments comprising diverse grid environments. Our algorithm was implemented in Matlab 2020a and evaluations considered the environments with grid maps with $50 \times 50$ cells and obstacles comprising convex and non-convex geometry a shown by Fig. \ref{maps}. Our computing environment was as follows Intel Core i7 @3.6GHz, 16 GB RAM. The time to generate lattice paths for a fixed $\alpha$ was in the order of 0.01 seconds, implying the fast performance for embedded platforms.

\begin{figure*}[ht!]
	\centering
	\includegraphics[width=0.98\textwidth]{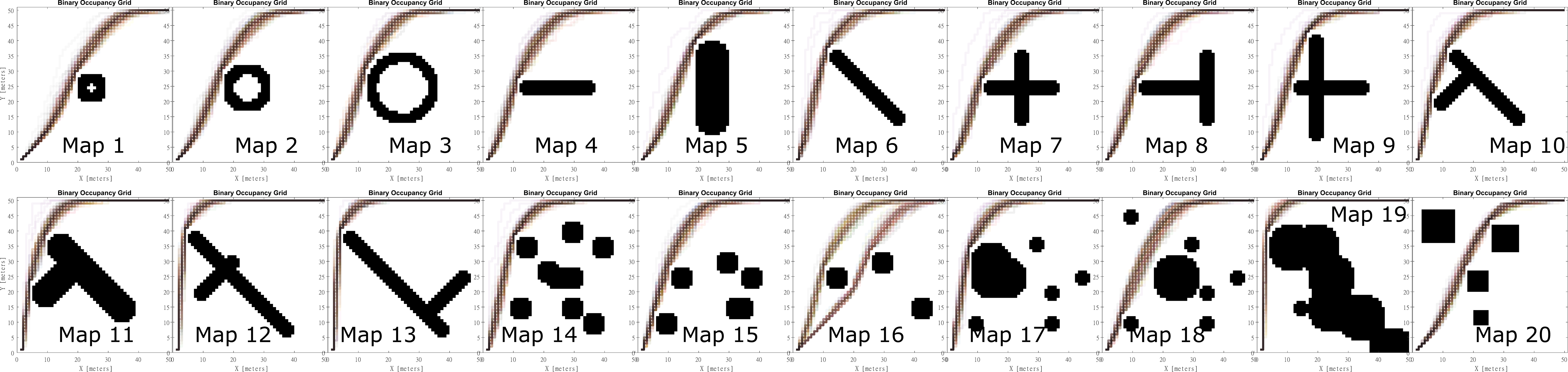}
	\caption{Obtained paths in each grid map environment over 20 independent runs.}
	\label{allpaths}
\end{figure*}

In order to show the performance frontiers of our approach, we used the following swarm heuristics to find optimal values of $\alpha$: Particle Swarm Optimization (PSO)\cite{pso95}, Particle Swarm Optimization with Speciation (PSOSP)\cite{psosp}, Differential Particle Scheme (DPS)\cite{dps21}, Particle Swarm with Fitness Euclidean Ratio (PSOFER)\cite{fer}, Differential Evolution with BEST/1/BIN mutation (DEBEST)\cite{de97}, Differential Evolution with RAND/1/BIN mutation (DERAND)\cite{de97}, Differential Evolution with Similarity-based Mutation Vector (DESIM)\cite{desim}, Strategy Adaptation Differential Evolution (SADE)\cite{sade}, Differential Evolution With Underestimation-Based Multimutation Strategy (UMSSADE)\cite{umssade19} Differential Evolution with Rank-based Mutation (RBDE)\cite{rbde}. For simplicity and without loss of generality, we set to obtain obstacle-avoiding lattice paths above the diagonal with origin at $(0, 0)$ and destination at $(n, n)$. Our motivation of using the above algorithmic set is to include distinct mechanisms of selection pressure, multimodality and trade-off mechanisms in exploration-exploitation. For each scenario and each algorithm, the objective function $F$ is the Euclidean distance of the path from the origin (bottom-left) to the destination (top-right) in the $50 \times 50$ grid map. For evaluations, we used 20 independent runs with 1000 function evaluations at the maximum, which allows to compare the behaviour over independent random initializations under tight computational budgets. Parameters for PSO-based schemes involve $\omega$  =  0.7, $c_1$  =   2.05, $c_2$  =  2.05, population size was set as 10. As for Differential Evolution algorithms, the crossover rate was $CR = 0.5$, the scaling factor $F = 0.7$. The coefficient $\beta$ involved in the Whitley distribution scheme in Rank-Based Differential Evolution (RBDE) was set as $\beta = 2$. The fine tuning of the afore-mentioned variables is out of the scope of this study. Other parameters used the default settings in the respective references.

\begin{figure}[t!]
	\centering
	\includegraphics[width=0.998\columnwidth]{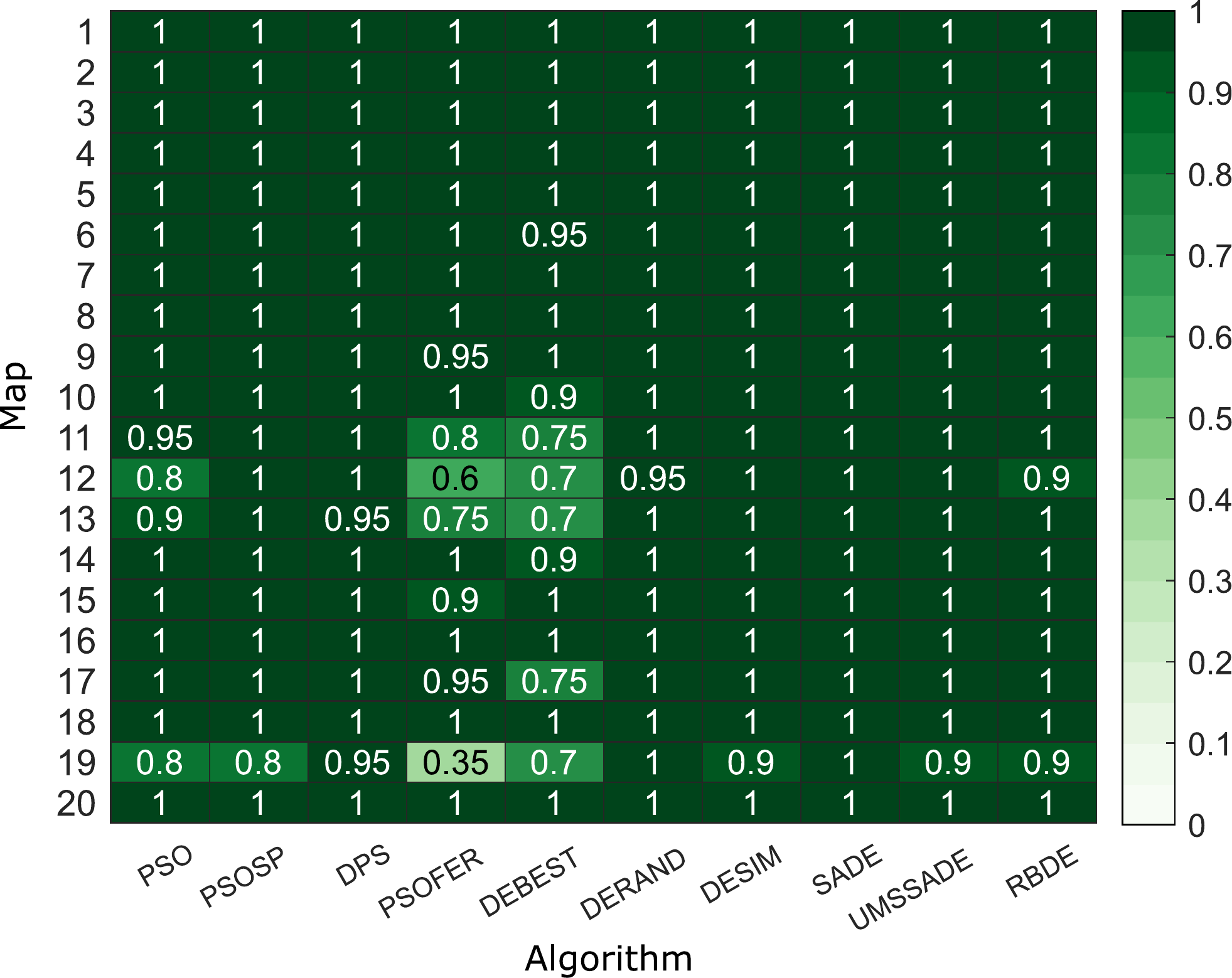}
	\caption{Success ratio of each algorithm over 20 independent runs.}
	\label{suc}
\end{figure}

To show the overall performance of all evaluated algorithms, Fig. \ref{allpaths} shows the obtained obstacle-avoiding lattice paths by all algorithms in all environments and all independent runs. Here, lattice paths are rendered with a transparency factor, thus lattice paths with darker colors imply the most common navigable regions. Also, by looking at Fig. \ref{allpaths} one can note that it is possible to obtain multimodal lattice paths over independent runs such as the case of Map 6.

In order to show the effectiveness of each swarm optimization algorithm, Fig. \ref{suc} shows the ratio of success at finding obstacle-avoiding lattice paths overall independent runs. For instance, by observing Fig. \ref{suc}, the ratio of 1 (0.8) denotes that the algorithm was able to find obstacle-avoiding lattice paths in 100\% (80\%) of the independent runs. Among the studied optimization algorithms, we can observe from Fig. \ref{suc} that only SADE was able to generate obstacle-avoiding lattice paths in all independent runs, and that maps 11, 12, 13 and 19 were the most challenging environments for most algorithms. We can also observe that the heuristics with exploitation features such as PSOFER and DEBEST are outperformed by heuristics having exploration and diversity inducing mechanisms such as SADE, DERAND, DESIM and DPS. Furthermore, most algorithms were able to find obstacle-avoiding lattice paths in the first ten environments, implying that the convex/nonvex nature of the obstacles has no significant effect on the performance of the algorithm. On the other hand, maps with narrow passages such as maps 11, 12, 13 and 19 were challenging for most optimization algorithms.

\begin{figure*}[ht!]
	\centering
	\includegraphics[width=0.98\textwidth]{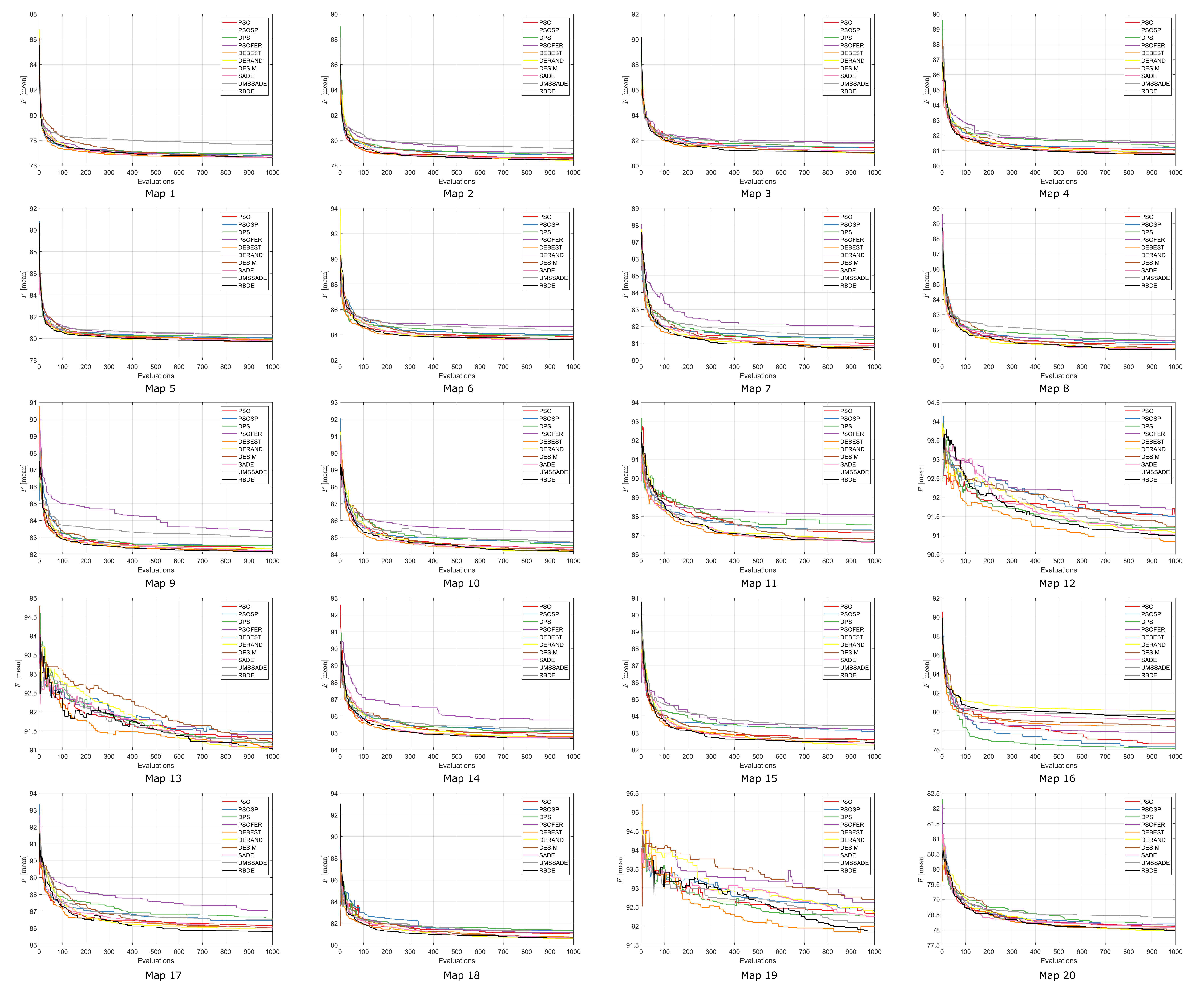}
	\caption{Mean convergence of the evaluated algorithms over 20 independent runs.}
	\label{conv}
\end{figure*}

To show the convergence characteristics of the studied algorithms, Fig. \ref{conv} shows the mean convergence behaviour over independent runs. By observing Fig. \ref{allpaths} and Fig. \ref{conv}, we can note the feasibility to compute obstacle-avoiding lattice paths in challenging navigation scenarios with reasonable number of function evaluations. The noisy behaviour of some convergence figures at Fig. \ref{conv}, e.g. at Map 12, 13, and 19 is due to some swarm optimization algorithms being unable to find obstacle-avoiding lattice paths in some independent runs (variability in the averaging over independent runs).

\begin{figure*}[ht!]
	\centering
	\includegraphics[width=0.98\textwidth]{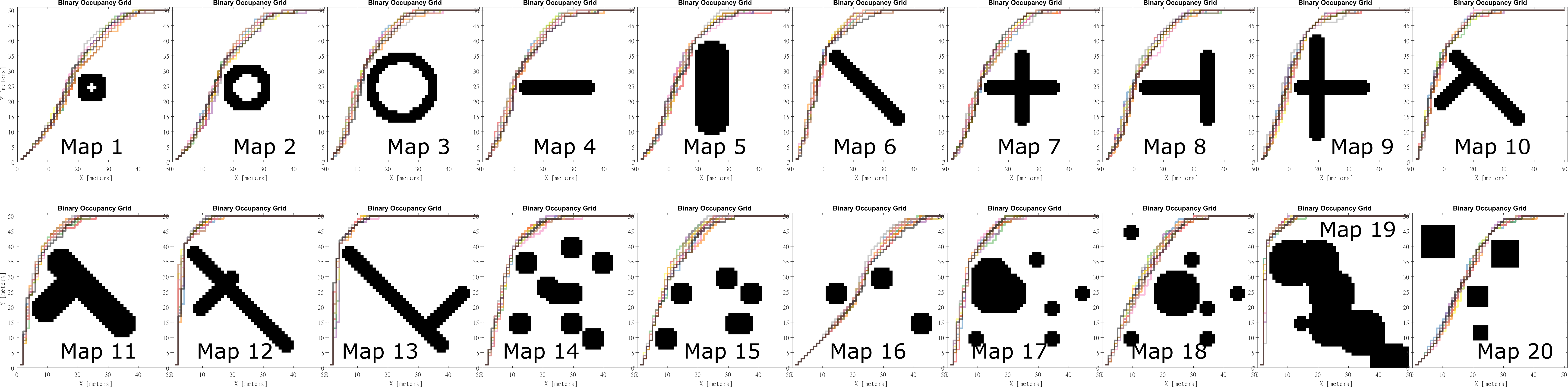}
	\caption{Best obtained paths in each grid map environment over 20 independent runs.}
	\label{allbestpaths}
\end{figure*}

\begin{figure*}[ht!]
	\centering
	\includegraphics[width=0.98\textwidth]{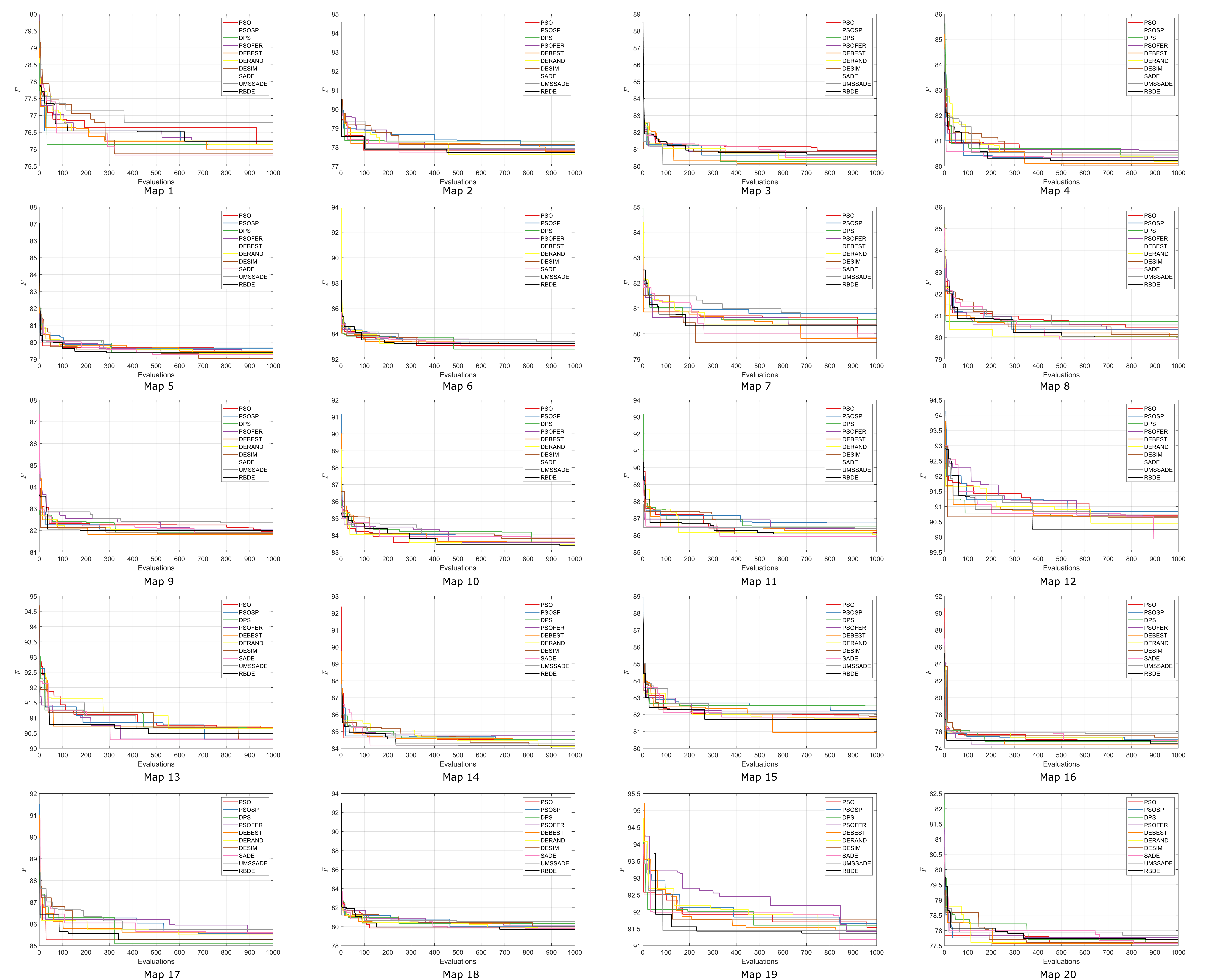}
	\caption{Lower bound of the convergence of the evaluated algorithms over 20 independent runs.}
	\label{minconv}
\end{figure*}

\begin{figure*}[t!]
	\centering
	\includegraphics[width=0.98\textwidth]{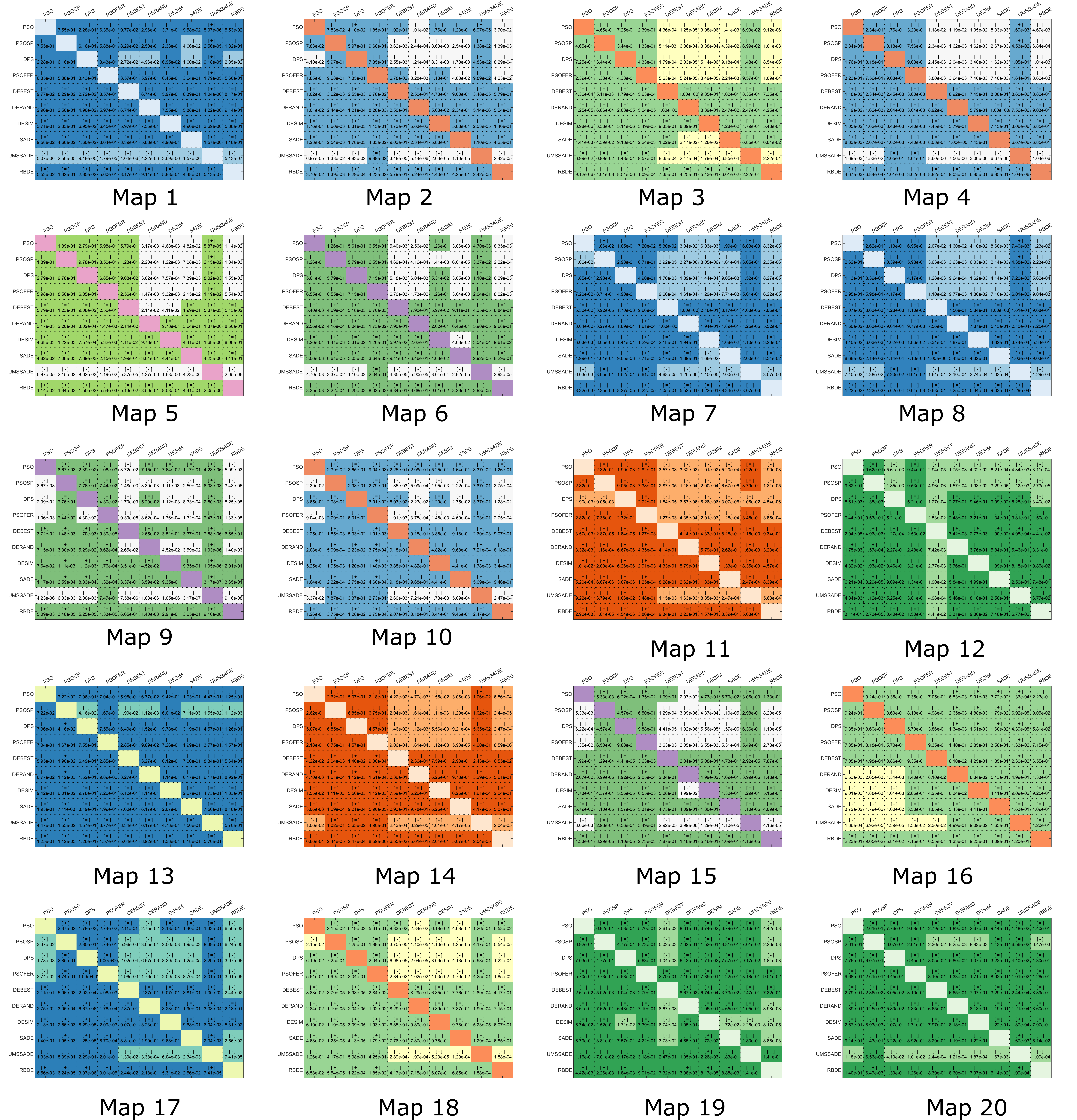}
	\caption{Statistical significance as measured by the Wilcoxon rank sum test in all maps.}
	\label{ptest}
\end{figure*}

On the other hand, Fig. \ref{allbestpaths} shows the best (shortest) path obtained by each optimization algorithm over all independent runs, and Fig. \ref{minconv} shows the lower bound of the convergence over independent runs. By observing Fig. \ref{allbestpaths} and Fig. \ref{minconv}, one can note that it is possible to obtain obstacle-avoiding lattice paths with similar topology, and that convergence occurs irrespective of the nonlinear stochastic algorithm over independent runs. It is also possible to study the statistical difference between the studied algorithms as shown by Fig. \ref{ptest}, which shows that explorative strategies are consistent in finding shortest lattice paths over all environments. The above-mentioned observations highlight the importance of exploration rather than exploitation when tackling the lattice path-finding problem. Investigating the fitness landscape for diverse geometries and large $n$ is left to future work.

The above-mentioned results show the feasibility and efficiency to generate collision-free lattice paths in grid maps. Due to one dimensional search problem and the 1-1 bijective property to ordered trees, our approach is potential to sample other combinatorial objects such as legal sequences of $n$ pairs of parentheses, triangulated $n$-gons, and other combinatorial objects based on catalan numbers. In future work, we aim at studying the smoothness considerations in paths\cite{gecco21,acc20}, the integration with trajectory tracking\cite{ccfl20}, the online adaptation and integration with other intelligent schemes such as Fuzzy Logic\cite{sii21}, the performance for very large $n$ and their further applications in combinatorial optimization in Robotics and Operations Research. We believe the proposed approach may find its use in planning and combinatorial optimization problems.

\section{Conclusion}

In this paper, we have proposed a new approach to generate obstacle avoiding lattice paths based on the 1-1 bijection to ordered trees, rendering a one dimensional optimization problem. Our computational studies using relevant swarm-based optimization heuristics has shown the feasibility to generate obstacle-avoiding lattice paths in challenging navigation scenarios consisting both of convex and non-convex obstacle geometries. Furthermore, our observations suggest that the explorative search strategies are effective over independent runs. In future work, we aim at studying further combinatorial optimization problems in Robotics and Operations Research; for instance the combinatorial problems involving binary trees with $n$ external nodes, the legal sequences of $n$ pairs of parentheses in modularity formation, and the triangulated $n$-gons in folding problems.




\bibliographystyle{IEEEtran}
\bibliography{mybiblio}



%

\end{document}